# *Soft-bubble*: A highly compliant dense geometry tactile sensor for robot manipulation

Alex Alspach[1], Kunimatsu Hashimoto[1], Naveen Kuppuswamy[1], and Russ Tedrake[1,2]

*Abstract*— Incorporating effective tactile sensing and mechanical compliance is key towards enabling robust and safe operation of robots in unknown, uncertain and cluttered environments. Towards realizing this goal, we present a lightweight, easy-to-build, highly compliant dense geometry sensor and end effector that comprises an inflated latex membrane with a depth sensor behind it. We present the motivations and the hardware design for this *Soft-bubble* and demonstrate its capabilities through example tasks including tactile-object classification, pose estimation and tracking, and nonprehensile object manipulation. We also present initial experiments to show the importance of high-resolution geometry sensing for tactile tasks and discuss applications in robust manipulation.

## I. INTRODUCTION

As we work towards deploying robotic systems to assist with everyday tasks, we must ensure that these robots are able to safely make contact with people, other robots and our environment. Unlike industrial robots that can operate with certainty about their tasks and surroundings, robots designed for our homes and other unstructured environments must be able to cope with large imprecision in their knowledge of the surrounding environment. In particular, for manipulation tasks, the ability to compensate for large uncertainty through touching and feeling is increasingly perceived as the solution to coping with the perceptual challenges posed by domestic environments. These challenges include clutter, occlusions, variable lighting conditions, and never before seen objects [1]. Due to their ability to directly capture interactions at the contacting surface, tactile sensors have the potential to be predominant when vision and other exteroceptive modalities are occluded [2] or incapable of sensing due to lack of sufficiently salient visual features [3]. Tactile sensing, coupled with the compliance needed to safely and robustly bump into and grasp objects in the environment, could prove hugely beneficial in hastening the deployment of robots at home. Beyond these design requirements, for commercially viable home robots, it is also necessary to consider component cost, weight, and manufacturability.

It is well recognized that mechanical compliance is a critical element in enabling robots to cope with unforeseen contacts [4] and offers greater intrinsic robustness to uncertainty [5]. One successful strategy for endowing robots with the requisite compliance is the use of elastomers and fluids as building materials [6]. For slow speeds and low masses, the passive compliance offered by these materials reduces the impulse of a collision while a deformable contact patch spreads force out over a larger area [7]. Elastomers have made their way into and onto traditionally rigid robots, including fully compliant air-powered sensorless grippers [8], [9] where precision is not required and compliance alone provides a stable grasp. However, constrained home manipulation tasks generally require more precision and care.

[1] Toyota Research Institute, One Kendall Square, Building 600, Cambridge, MA 02139, USA, alex.alspach@tri.global, kunimatsu.hashimoto@tri.global, naveen.kuppuswamy@tri.global, russ.tedrake@tri.global
[2] The Computer Science and Artificial Intelligence Laboratory, Massachusetts Institute of Technology, 32 Vassar St., Cambridge, MA 02139, USA, russt@mit.edu

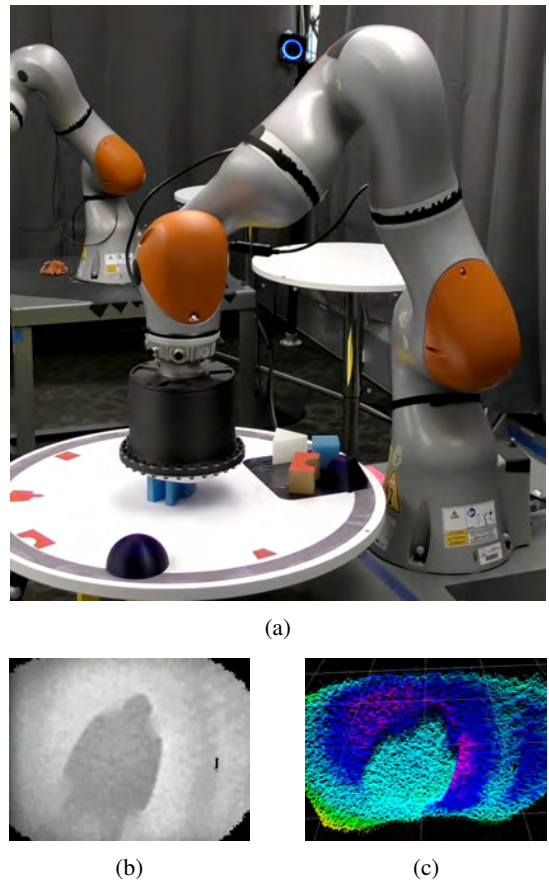

Fig. 1: (a) The *Soft-bubble*, a high-resolution tactile sensor, is attached to a KUKA iiwa and pressed against a robot-shaped block, allowing classification and nonprehensile manipulation (a). The resulting depth image (b) and point cloud (c) are shown as well.

In addressing the need to sense and manage physical interactions, there are several themes that have been tackled so far: sensing accidental contacts and reacting suitably [10], tracking and controlling contact forces during intentional contact [11], tactile exploration for objects that are hard-to-see or occluded by clutter [3], classification of object type and shape inference [12] and sensing the quality of induced grasps [13]. For tactile sensing of object class (shape, material, etc.) and object state (pose, velocity, etc.), contact geometry sensing enables an understanding of surface and other physical properties [14] as well as pose refinement in order to manipulate the object accurately [15]. Despite advancements on these themes, the fundamental difficulties and open questions in modeling compliant contact mechanics [16] have limited the adoption and deployment of soft tactile sensors. Although data-driven methods have been employed as attempts to overcome the modeling difficulties [13], [17], there remains a lack of highly compliant mechanisms which also incorporate high-resolution contact sensing.

In this paper, we present our *Soft-Bubble*, pictured in Fig. 1, a new kind of tactile sensor and end effector that combines the advantages of a highly compliant elastomeric structure with the ability to sense the detailed geometric features of contacting objects. The proposed sensor captures deformation of a thin, flexible air-filled membrane using an off-the-shelf depth sensor. This sensor is built using accessible fabrication methods and is composed primarily of off-the-shelf components and materials. The resulting sensor is highly compliant, lightweight, robust to continued contact, and outputs a high-resolution depth image that is ideal for manipulation applications. We demonstrate the efficacy of these features through three case studies: (a) object shape and texture classification using a deep neural network, (b) an object sorting task using the *Soft-bubble* end effector for nonprehensile manipulation, and (c) object pose estimation and tracking. For the latter two demonstrations, an instance of the proposed sensor is mounted on a KUKA iiwa arm and used to classify, explore and manipulate blocks of various shapes and sizes, thus demonstrating the dual application as a tactile sensor and capable end effector.

This paper is organized as follows: Section II presents a background of dense geometry and compliant tactile sensing, Section III illustrates the system design and IV presents the results of experimental evaluation. Finally, we discuss our findings and outlook for future work in Section V.

## II. RELATED WORK

In this section, we will briefly review some of the technologies and algorithms most relevant to the *Soft-bubble*. The field of tactile sensing has seen dramatic growth in the last 25 years [18]. A wide variety of technologies have been proposed to sense shape, texture, hardness, temperature, vibration or contact forces [19], but open questions still remain as to how tactile sensing can be employed with the explicit goal of improving a robot's ability to manipulate the world around it. How high a spatial resolution is necessary for tactile sensing? Is either geometry or force sensing more important than the other?

High-resolution tactile sensors, such as GelSight [14], GelSlim [20] and FingerVision [21], use cameras to gather large amounts of data over relatively small contact areas. GelSight in particular uses precise internal lighting and photometric stereo algorithms to generate height maps of contacting geometry. This 3D information can be used in a model-based framework [15] wherein it can be fused with external sensors, or run through particle filters [22] that capture the complicated contact mechanics or used to directly sample the contact manifold as dictated by the manipuland and gripper geometries [23]. Alternative data-driven approaches also exist [13], although these methods have been largely employed in the domains of slip detection, grasp stability identification [17] and material property discrimination.

The *Soft-bubble* draws influence from these camera-based tactile sensors, particularly on its use of an off-the-shelf depth camera and an opaque membrane which drapes sensed object surfaces in consistent color and reflectance properties. Mechanically, the *Soft-bubble* is able to deform around a contacting object more freely and drastically than the gel-based sensors above, potentially making a larger slice of the object's geometric form available to the sensor. As a result of using a self-contained depth sensor, precisely placed illumination and 3D reconstruction algorithms are not needed to capture deformation. This allows our technology to work on a large range of free-form membrane shapes. The fabrication process is cheap, simple and repeatable. The use of air over gel also simplifies fabrication and keeps the sensor lightweight, making it a suitable sensor for low-cost, low-payload robots. Employing the resilience of latex, the sensor membrane can withstand rough treatment while worn components are easily replaced. Finally, the compliant, high-friction membrane surface offers large contact patches and form closure via deformation around an object, making this sensor well-suited for contact-heavy manipulation tasks.

## III. HARDWARE DESIGN

The *Soft-bubble* dense geometry sensor design consists of three main functional components: an elastic membrane sensing surface, an airtight hull that allows pressurization of the membrane, and an internal depth sensor, as shown in the dimensioned cross-sectional and exploded views in Fig. 2. The depth sensor is located inside of the airtight hull pointed at the interior surface of the membrane to measure contact induced deformation. Inflated, the sensing membrane forms a compliant spherical cap. The membrane is inflated to a height of 20-75mm. The stiffness of the bubble increases with height/internal pressure. The membrane material is laser-cut from 0.4mm thick latex sheet. Thinner or thicker material can be used with trade-offs between sensitivity and durability. For the experiments presented in Section IV, the sensing membrane is inflated to a height of 50mm with an internal air pressure of roughly 0.25psi.

### A. Internal depth camera

The depth camera in the embodiment presented here is the PMD CamBoard pico flexx [24] time-of-flight (ToF)

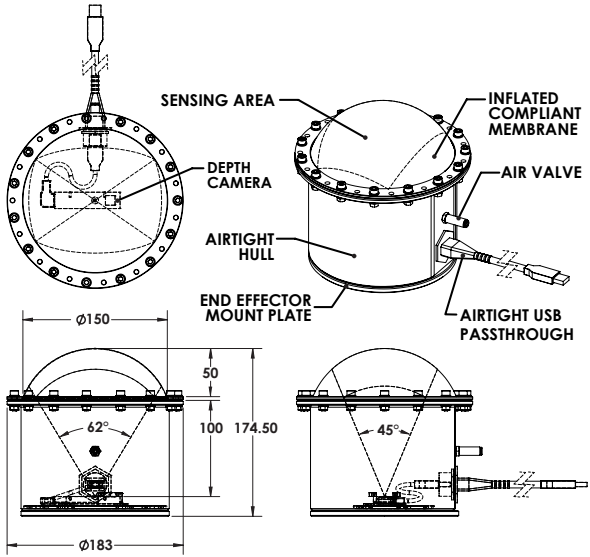

Fig. 2: Dimensioned sensor assembly. The size is based on the specifications of the chosen depth sensor, the PMD pico flexx. All dimensions in $mm$.

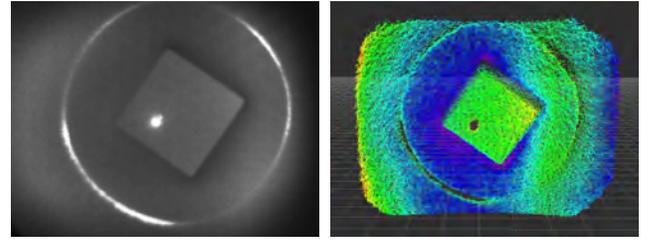

(a) IR image     (b) Point cloud

Fig. 3: Glare due to excess emitter intensity can be seen in the form of bright spots in the IR image and absent points in the point cloud.

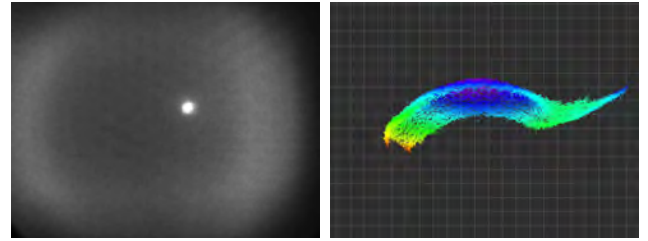

(a) IR image     (b) Point cloud

Fig. 4: The dimly illuminated area to the right of the IR image forms during close-range imaging as a result of emitter and imager FOV offset. The uncompensated, insufficient illumination results in points with depths represented as further than they actually are, as seen on the right edge of the point cloud.

camera. This depth camera was chosen for its small size (68×17×7.35mm), USB connectivity and short specified minimum sensing range (100mm). The sensor has a resolution of 224×171 pixels and can provide depth images at up to 45 fps. At close range, the specified depth resolution of this sensor is $\leq 2\%$ of the actual distance. The sensor weighs 8 grams and has a field of view (FOV) of 62°×45°. When inflated to 50mm, the spacial sensing resolution on the membrane is roughly 2 pixels/mm$^2$

The pico flexx depth camera is designed to sense over the range of 0.1-4 m. The optics are focused and the sensor is calibrated to work best in the middle of that range. When operating near the minimum range in a confined space, two issues become evident: the IR emitter is too bright and the effect of the IR emitter-imager offset, negligible at further distances, becomes non-negligible. The emitter produces bright spots on regions of the inner membrane surface causing sensor saturation seen in Fig. 3. Also, at this range, the offset between the emitter and the imager FOVs leaves a poorly illuminated region where the FOVs fail to overlap. This asymmetric dark region in the IR image reveals itself in depth measurements as further away than it actually is, seen in Fig. 4.

*B. Airtight hull*

The hull structure provides mounting points for the depth sensor, elastic membrane, a sealed USB 3.0 passthrough and an air valve. This structure also provides mounting points for attaching the tactile sensor assembly to a robot. The dimensions of the hull and membrane are dependent on the chosen depth sensor's range and FOV specifications. At the depth sensor's minimum sensing distance, 100mm, the area covered by the FOV is 99.5cm$^2$. The diameter of the membrane is chosen so that the entire FOV falls onto the membrane. At 50mm inflation, this provides a sensing surface area of 175.4cm$^2$. The dome's total surface area is 261.4cm$^2$, which leaves about 86cm$^2$ out of the depth sensor's FOV.

Towards the goals of low overall cost and weight, as well as accessible fabrication, the airtight hull is 3D printed. This particular version is printed on a Markforged X7 fused deposition modeling (FDM) printer [25]. FDM printing generally results in a porous part that can not hold air, therefore the internal surfaces of the hull, passthrough holes, and the O-ring channel where the latex membrane meets the hull are painted with two or more coats of Rust-Oleum 265495 Leak Seal Flexible Rubber Sealant.

The hull contains internal mount points for the depth sensor. The USB passthrough used in this assembly is a sealed Molex 84733-series USB 3.0 Type A connector. The valve installed on the hull for pressurization and depressurization of the sensor is a M5 threaded push-to-connect fitting (McMaster-Carr part number 1201N11) for 4mm OD tubing that automatically shuts when the tube is removed.

The complete sensor is assembled by installing the USB passthrough and air valve, bolting the depth sensor to the interior floor of the hull, then connecting the sensor to the USB passthrough. An O-ring is placed into the channel on top of the hull flange. The circular latex membrane is glued to a 3D printed frame while flat so that no stretching or sagging occurs while assembling. This frame and membrane subassembly is then bolted to the hull flange so that the O-ring seals between the hull's rubberized O-ring channel below and the latex membrane above. The complete assembly can then be attached to the end effector of a robot. The assembled *Soft-bubble* tactile sensor can be seen in Fig. 1a installed on a KUKA iiwa.

*C. Parallel gripper concepts*

As depth sensors and their sensing ranges become smaller, this technology will be incorporated into sensors of a scale more suitable for standard robot grippers. Figure 5 shows a sensorless bubble-based parallel gripper prototype able to execute robust and versatile grasps on arbitrary objects due to its compliant, high-friction latex membrane. Research and development towards sensorizing these smaller bubbles in the same way as the *Soft-bubble* is ongoing, but the benefits of highly compliant grasps are immediate.

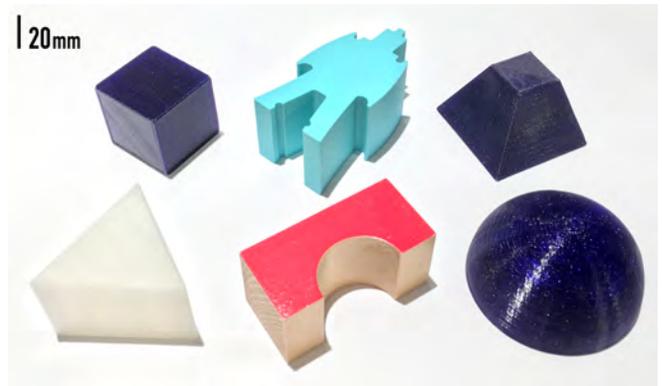

Fig. 6: *Six objects* used for the tactile object classification experiment. For scale, the cube is 40×40×40mm.

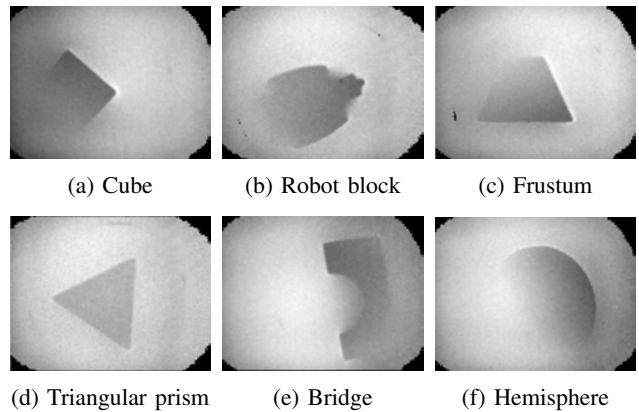

(a) Cube  (b) Robot block  (c) Frustum

(d) Triangular prism  (e) Bridge  (f) Hemisphere

Fig. 7: Tactile depth images of the *six objects*, captured by our *Soft-bubble* tactile sensor.

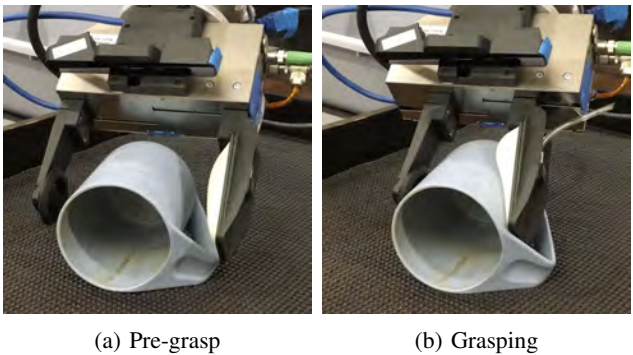

(a) Pre-grasp  (b) Grasping

Fig. 5: Sensorless *bubble*-based gripper prototype able to robustly grasp due to its compliant, high-friction surface. This gripper-finger prototype lends insights towards specifications for small, short-range depth sensors.

## IV. Experiments

For an initial demonstration of the *Soft-bubble* tactile sensor capabilities and potential applications, we conducted multiple experiments: two tactile-based object classification experiments, an object sorting manipulation experiment, as well as tactile pose estimation and tracking experiments. Overall, these experiments show the utility of high-resolution contact geometry sensing and highly compliant manipulation. A video of these experiments can be found here: *https://youtu.be/sDfNkJzZ7RY*

*A. Classification*

We use ResNet18 [26], a state-of-the-art deep neural network, as the object classifier for both of the classification experiments described in this section. As input, the network takes a depth image obtained when the *Soft-bubble* sensor is pressed up against an object. The network outputs the probabilities for each object class.

First, we explored whether or not an object classification task could be performed using the data output by the *Soft-bubble*. We set up seven classes for this experiment: six unique objects and an extra class called "no-touch" for when the *Soft-bubble* was free from contact. The objects chosen for this first experiment are distinctly shaped blocks (Fig. 6).

To gather data, we recorded depth image streams from the *Soft-bubble* while each of the *six objects* were pressed into the bubble's membrane while varying contact location, object orientation, and contact force (Fig. 7). We aimed to record contact geometry data for a wide range of poses, including partial views of local features, as illustrated by the selection of robot block depth images in Fig. 8. Depth image streams were also recorded for the "no-touch" case. The various data was collected using two separate *Soft-bubble* assemblies with slight differences between them. The internal pressures of the bubbles were varied throughout to further diversify the data collected. These depth data streams for contacting objects were then passed to a filtering algorithm which discarded "no-touch" frames based-on a depth deviation threshold before sampling. For the "no-touch" stream, we simply sampled from all captured frames. Consequently, we collected 1,000 training images and 200 validation images for each of the seven classes. To train the network, cross-entropy loss and stochastic gradient descent were used as the loss function and optimizer, respectively. We used 0.1 for the learning rate with a 0.1 decay every 30 epochs, 0.9 for momentum, $10^{-4}$ for weight decay, and 32 as the mini-batch size. Note that the network was trained from scratch since publicly available pre-trained weights are usually trained using RGB images, differing from our choice of network input. The network was trained up to 80 epochs and achieved 98.14% average top-1 accuracy as its best performance throughout the training on our validation data. This result indicates that the *Soft-bubble* can be successfully used for practical object classification tasks.

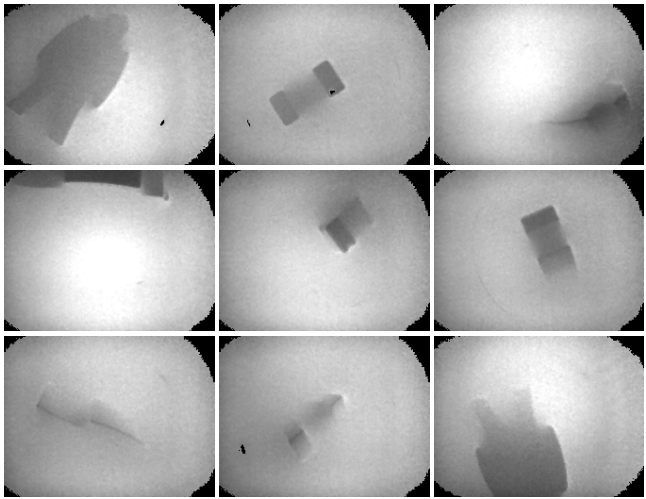

Fig. 8: Tactile depth images of all *six objects* were recorded for a wide range of object poses, including partial views, as illustrated by the selection of robot block tactile depth images above.

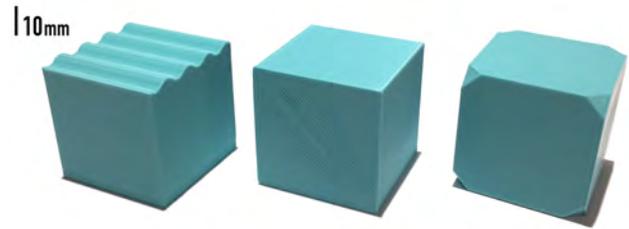

Fig. 9: *Three cubes* including a standard cube (middle), wavy face (left) and cut-off corners (right) used in classification experiment exploring the resolution necessary for discriminating based on small (relative to resolution) geometric features. For scale, the standard cube is 40×40×40mm.

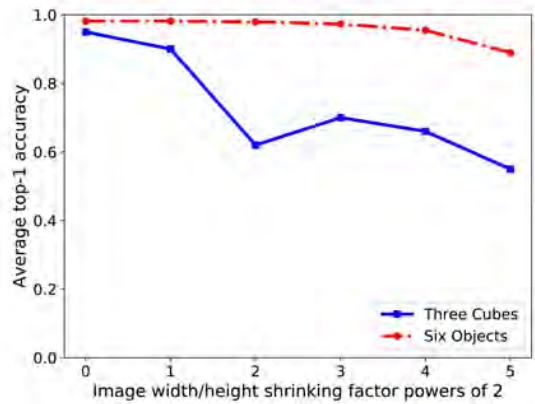

Fig. 10: Average Top-1 classification accuracy for experiments with varying image resolution using the *six object* and *three cube* datasets.

Next, we investigated the importance of input image resolution. To scale this tactile sensor down to a size usable on human hand-scale gripper hardware (concept design shown in Fig. 5), a smaller short-range depth camera must be developed. The exploration into the necessary minimum spatial resolution for given tactile tasks will help determine target specifications for hardware development. In this experiment, we used two datasets; the previously detailed *six object* dataset, and another one called the *three cube* dataset which consists of four classes: "no-touch" and three types of cubic objects, congruent in size, but with unique surface features (Fig. 9). These cubic objects were chosen so that differences in shape between objects are subtle, especially as input image resolution is decreased. The *three cube* dataset contains 1,000 training images and 500 validation images per category, gathered via the previously mentioned procedure. The same network architecture used before is again used for the *six object* and *three cube* datasets. The network is trained separately from scratch on the *six object* and *three cube* datasets. The training is repeated six times for each dataset, each with varying image resolutions; the width and height of the input image is first multiplied by $2^{-N}$, where $N$ is a hyperparameter which varies from 0 to 5 over each training cycle. Each image is then resized to 224×224 pixels to fit the input size of our network. We measured average top-1 accuracy as the metric for this experiment. The results are shown in Fig. 10. This illustrates that the accuracy is reasonably good at higher resolutions but drops as input image resolution decreases. This tendency is especially evident on the *three cube* dataset where the differences between each object are more subtle than those of the *six object* dataset. The accuracy for the *three cubes* with varying surface features drops significantly at $N = 2$. The point clouds corresponding to the original resolution and quartered resolution where $N = 2$ are shown in Figs. 11 and 12, respectively. From these results, we can argue that the higher the resolution, the better the accuracy will be for classification tasks. The results also suggest that tactile resolution must be carefully chosen based on task demands.

### B. Pose Estimation and Tracking

As a third exploration into the capabilities of the *Soft-bubble*, we explore pose estimation and tracking of objects with known geometries using the depth image captured by the sensor. To estimate the pose, we used the well known Iterative Closest Point (ICP) algorithm [27].

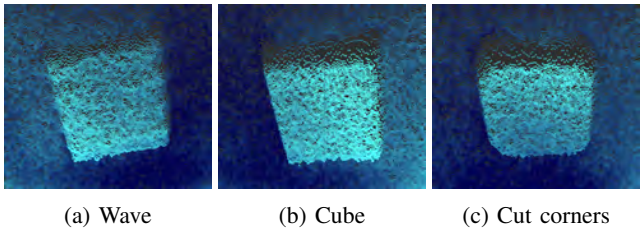

(a) Wave      (b) Cube      (c) Cut corners

Fig. 11: Full resolution point clouds of the *three cubes* used in varying image resolution classification experiment.

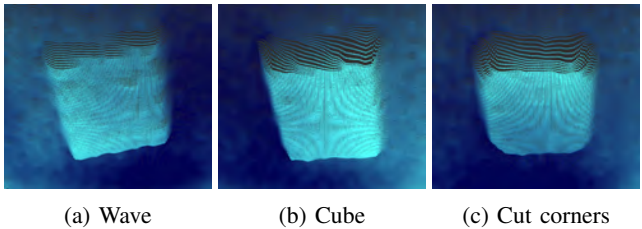

(a) Wave      (b) Cube      (c) Cut corners

Fig. 12: Lower resolution (length and width divided by $2^2$) point clouds of the *three cubes* used in varying image resolution classification experiment. It is at this resolution that a drop-off in accuracy can be seen in Fig. 10.

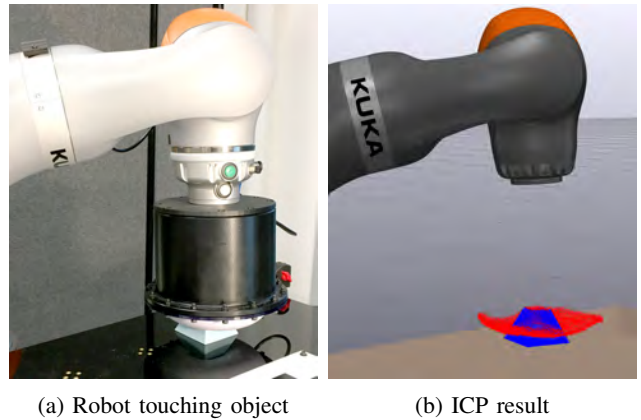

(a) Robot touching object      (b) ICP result

Fig. 13: ICP-based pose estimation of an object with known geometry using the *Soft-bubble* tactile sensor. Estimated pose of a pyramidal frustum rendered in blue with the point cloud from the *Soft-bubble* in red.

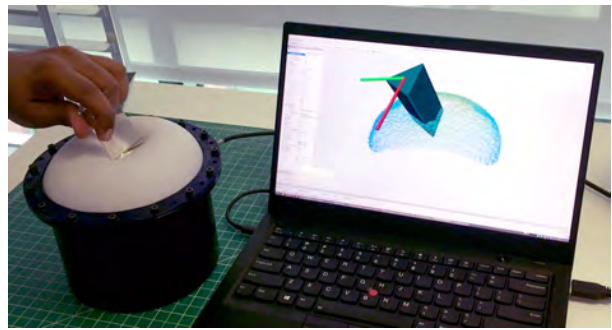

Fig. 14: ICP-based pose tracking of a transparent glass triangular prism of known dimensions. Poses of objects that are typically hard-to-see with non-tactile sensors can be tracked using our *Soft-bubble* tactile sensor.

The setup presented in Fig. 13a consists of the KUKA robot arm along with a table comprising mounting points for interchangeable target manipulands. The extrinsic location of the target board with respect to the World frame $W$ (assigned as the base of the robot) was obtained by attaching a checker board to the table and using external RGB camera-based extrinsic calibration. The pose of the pico flexx camera with respect to the distal link of the KUKA arm was obtained from CAD and was used for transforming the captured point clouds to the reference frame $W$. For the purposes of this experiment, the KUKA was commanded to execute various Cartesian trajectories to reach out and press onto an object from a set of 10 approach angles sampled uniformly over a cone of aperture $15°$ centered on the object. The target manipuland chosen was a pyramidal frustum, identical to that shown in Fig. 6. Each trajectory resulted in the object pushing into the bubble up to a depth 4cm. Over the course of the experiment, configurations of the robot and point clouds from the sensor were logged and used for pose-estimation.

The ICP implementation from the PCL library [28] was used along with a dense model of the target object to compute the pose. The optimal pose was obtained by executing ICP from 1-12 initial orientations about the small face of the frustum and choosing the best fit result. For each initial orientation, up to 25% of the model opposite the surface assumed to be in contact (i.e. the small face) was cropped off in order to capture only the surfaces visible to the *Soft-bubble*. This procedure eliminated potential false registration results. This simplification relies on having a reasonable local prior assumption of the pose, the bounds of which are dependent on the variant of ICP that is employed as well as the geometry of the reference model. The image in Fig. 13b shows the converged result for the pose of the pyramidal frustum depicted as the object model superimposed onto the measured point cloud. Computing the pose took approximately 0.5 seconds. It can clearly be seen that the pose of contacting objects with simple geometries can be computed using this sensor.

We extended this approach into a basic pose tracker that can operate at 1-2Hz. Figure 14 shows pose tracking of a glass prism that is pushed into the *Soft-bubble* by hand. The initial poses used for ICP are drawn from a moving window comprising two fixed rotations about the assumed contact face, as well as the orientation corresponding to the last successful ICP result. This results in relatively smooth tracking of poses when a target object is being moved or rotated on the sensor surface while maintaining contact.

More detailed analysis of the pose accuracy, effect of complicated geometries and sensor resolution are out of the scope of this paper.

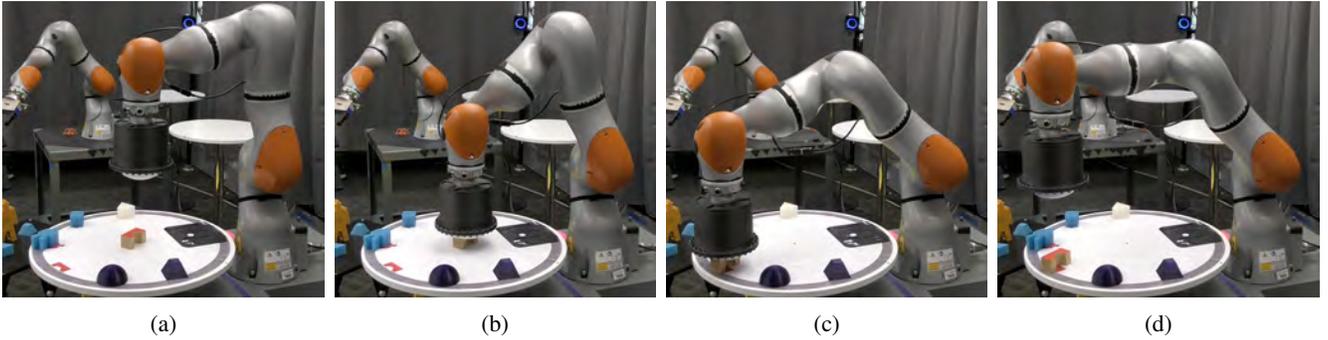

|  (a)  |  (b)  |  (c)  |  (d)  |

Fig. 15: The robot positions itself above an object (a), presses against and classifies the object (b), pushes it along the table to the appropriate location (c) then retracts to repeat (d). This demonstration of a simple but real-world sorting manipulation task illustrates the value of this hardware as a tactile sensor as well as a versatile end effector.

*C. Robot Object Sorting*

We integrated the *Soft-bubble* tactile sensor, the *six object* classifier from the first experiment and a robot to perform a simple but real-world task. The *Soft-bubble* is attached to the end of a KUKA iiwa robot arm, as shown in Fig. 1a. We put each of the *six objects* from the first experiment down on the center of a table in front of the robot. The task for the robot is to reach down, touch each object, and classify using tactile information. Once identified by the object classifier, the robot pushes the object toward a designated area on the table. All of the robot's individual motion primitives, i.e. pushing down, sliding objects and going back to the initial pose, are pre-scripted. While the classifier is running at all times, object classification for sorting is performed between the pushing down and sliding away motions. Fig. 15 illustrates the robot's execution of the sorting task. Through this integration, we demonstrate that this hardware is not simply a tactile sensor but, thanks to its highly compliant, robust construction, a capable and versatile end effector as well.

## V. DISCUSSION AND OUTLOOK

As mentioned in Section III, the large form factor of the tactile end effector presented here is dependent on the specifications of the depth sensor. While we used one of the smallest and most short-range-capable depth sensors on the market, a smaller depth sensor with a shorter sensing range and a wide FOV is needed in order to deploy the *Soft-bubble* on human hand-scale grippers (Fig. 5) and other space-limited parts of a robot. Smaller depth sensors also lead to interesting possibilities for large bubbles with multiple synchronized depth sensors producing a fused point cloud. Again, we envision future versions of this sensor located on various parts of a robot, not only the end effector.

As we aim to scale the *Soft-bubble* down, an understanding of the tactile resolution necessary for given tasks is needed to drive specifications for small, short-range depth sensors. In Section IV, we show that classification and pose estimation tasks can be performed using depth images from the *Soft-bubble*, as well as that image resolution affects the classification accuracy, especially when the differences in object shape or surface features are subtle with respect to sensor spatial and depth resolutions. We will continue to explore the effects of resolution with respect to tactile sensor size and use case.

The *Soft-bubble* currently senses geometry only, i.e. extracting contact forces requires additional modeling and analysis. Currently, a naive geometry-based method is used to determine which points on the sensor's membrane likely correspond to points on a contacting object. For convex objects, this method is susceptible to falsely identifying bridging between contacts due to membrane tension as part of a contacting object's surface. Future work includes static and dynamic modeling of these soft mechanics so that contact location and pressure on the membrane may be estimated based on sensed deformation and membrane physics. With the addition of dots or other trackable features on the inner surface of the membrane, shear forces and moments can be estimated as well. Modeling will also allow the sensor contact mechanics and output to be simulated. Methods for calibrating the tactile sensor's depth output, as well as for quantifying measurement error and sensor noise are actively under development.

As illustrated by the integrated sorting demo shown in Fig. 15, our manipulator with a highly compliant *Soft-bubble* tactile end effector is capable of both sensing and manipulating objects. Future research directions include tactile-based pose refinement, contact patch and contact pressure estimation, dexterous manipulation, contact-rich planning and control and exploratory techniques for locating and manipulating occluded and hard-to-see rigid and deformable objects.